\documentclass[conference,letterpaper]{IEEEtran}
\usepackage[T1]{fontenc}
\usepackage[utf8]{inputenc}
\usepackage{amsmath,amssymb}
\usepackage{booktabs}
\usepackage{graphicx}
\usepackage{placeins}
\usepackage{float}
\usepackage{xcolor}
\usepackage{colortbl}
\usepackage{cite}
\usepackage{balance}
\usepackage[hidelinks]{hyperref}
\graphicspath{{figures/}}
\definecolor{scoregreen}{HTML}{26734D}
\newcommand{\method}{\textsc{SCoRe}}

\newcommand{\one}{\mathbf{1}}

\newif\ifrevisionmarks
\DeclareRobustCommand{\rev}[1]{%
  \ifrevisionmarks\textcolor{red}{#1}\else#1\fi}
\newcommand{\revisionink}{\ifrevisionmarks\color{red}\fi}
\newenvironment{revision}{\par\begingroup\revisionink}{\par\endgroup}

\revisionmarksfalse
\begin{document}

\title{\method: Support-Calibrated Role Fusion for\\
\makebox[\textwidth][s]{Training-Free Cross-Domain Few-Shot Segmentation}}

\author{%
\IEEEauthorblockN{\rev{Abhishek Kathpal}}
\IEEEauthorblockA{\rev{Inception Robotics, LLC}\\
\rev{College Park, MD, USA}\\
\rev{\texttt{kathpal.abhishek@gmail.com}}}
\and
\IEEEauthorblockN{\rev{Adarsh Jagan Sathyamoorthy}}
\IEEEauthorblockA{\rev{Inception Robotics, LLC}\\
\rev{College Park, MD, USA}\\
\rev{\texttt{adarshjagan3895@gmail.com}}}
\and
\IEEEauthorblockN{\rev{Utsav Patel}}
\IEEEauthorblockA{\rev{Inception Robotics, LLC}\\
\rev{College Park, MD, USA}\\
\rev{\texttt{erutsavpatel22@gmail.com}}}
}
\maketitle

\begin{abstract}
We present \method \,(Support-Calibrated Role fusion), a framework to utilize two complementary vision foundation model-based (VFM) segmenters for training-free cross-domain few-shot segmentation (CD-FSS). Our approach is based on the observation that current VFMs for CD-FSS possess capabilities where one model locates the correct object, while another model computes the better contour. Our approach uses: 1. a method to identify a VFM segmenter to \textit{lead} Few-Shot Segmentation (FSS) task for a given concept, 2. a novel deterministic boundary correction method using bands of near- and far-pixels at the lead segmenter's predicted boundary of an object, and 3. a region correction mechanism for the other segmenter to conservatively prune components, fill holes, and bridge short gaps on the lead segmenter's predicted masks to improve accuracy.
\rev{We evaluate \method{} using RSRM and FSS-SAM3 across 3,314 matched
five-shot episodes from 10 datasets. \method{} achieves 72.738\% foreground
mIoU averaged across datasets, a gain of 5.210 percentage points over RSRM
(paired 95\% CI [4.618, 5.972]), with per-dataset gains up to 8.461 points
over the strongest standalone method and the best point estimate on 7 of
10 datasets.}
\end{abstract}


\section{Introduction}

\begin{figure*}[!t]
\centering
\includegraphics[width=.96\textwidth]{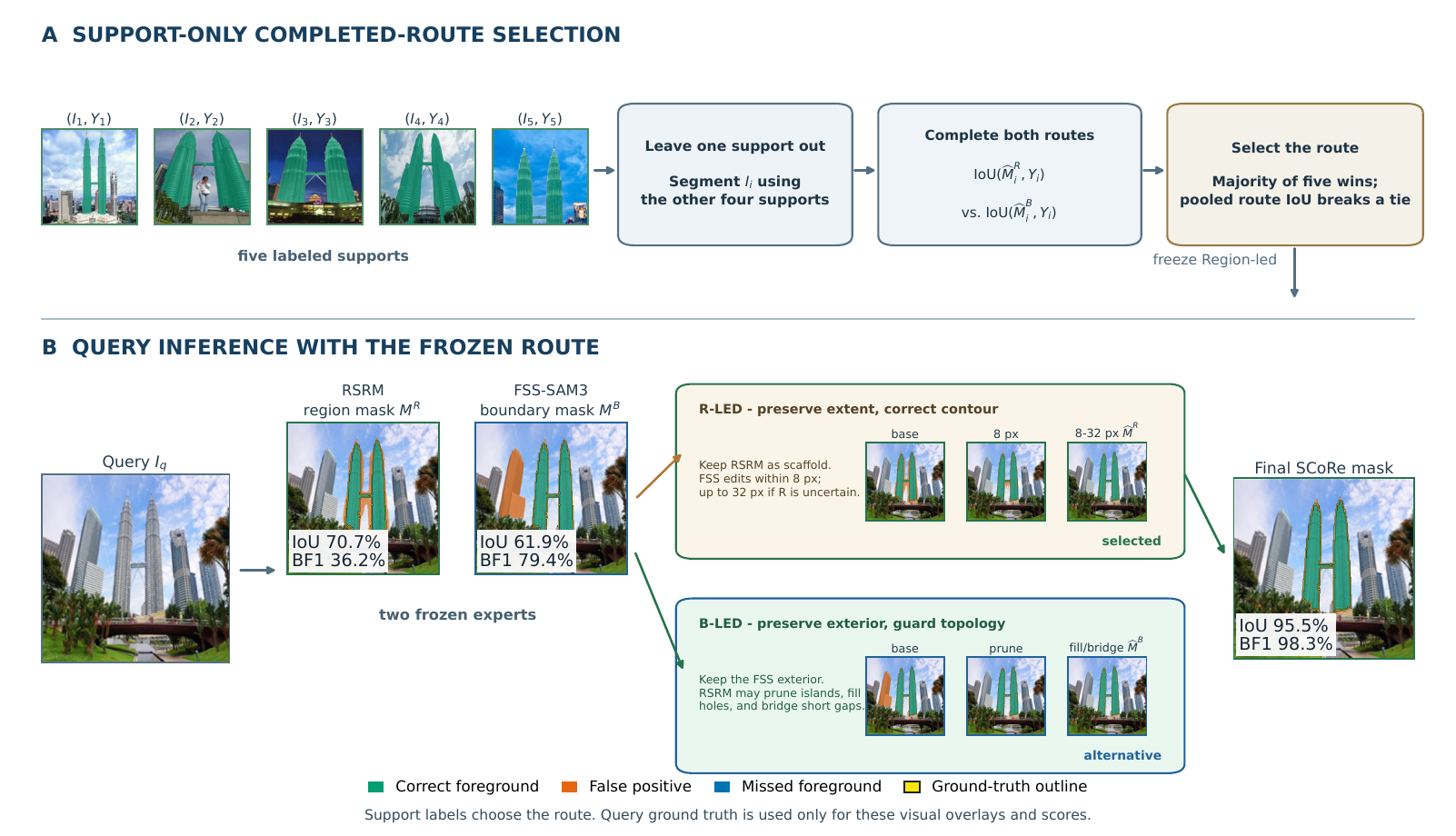}
% \caption{Overview of \method{}.  Five leave-one-support-out folds compare the
% two completed routes.  Their majority freezes one role assignment for the
% routing unit.  The selected route then combines the same two query predictions.
% Support labels are used only during calibration; query labels are used only for
% evaluation.}
\caption{Overview of \method{} and the complementary strengths of its
experts. (A) Five leave-one-support-out folds compare the two completed
routes; a majority vote selects and freezes the route for query inference.
(B) RSRM more accurately identifies the target region but produces
imprecise contours, whereas FSS-SAM3 provides more accurate boundaries
but includes false-positive regions (the extra building in orange). The selected region-led route
preserves RSRM's semantic extent and uses FSS-SAM3 to refine its contours,
producing a mask with substantially higher region overlap and boundary
accuracy in the illustrated example. Support labels are used only for
route calibration; query ground truth is used only for visualization
and evaluation.}
\label{fig:overview}
\end{figure*}

\rev{Cross-domain few-shot semantic segmentation (CD-FSS) predicts a
query mask for a target concept using $k$ \textit{support images} and
their segmentation masks.} The task is especially challenging because a model trained using certain kinds of images and concepts (say RGB images of natural objects) finds it difficult to understand the features of objects in completely different domains such as X-ray or satellite images. 
% Can add a sentence about methods using small transformer models.

Naturally, the latest Vision Foundation Models (VFMs) that have been trained in internet-scale data have demonstrated state-of-the-art performance in adapting to new domains due to their excellent class-agnostic correspondence capabilities. Recent methods like HERA \cite{hera} have proposed finetuning adapters for the underlying VFM such as DINO v3 per segmentation task. RSRM \cite{rsrm}, FSS-SAM \cite{fsssam3} have proposed training-free approaches by just using the vision encoders of the underlying DINO v3 and SAM VFMs. 

Empirically, we observe that DINO and SAM-based methods have complementary capabilities. DINO-based methods demonstrate strong object localization with imprecise contours, while SAM-based methods compute more accurate contours but on an incorrect object. Further, their performance varies significantly in different domains, which stems from their training. DINO/DINOv3 is trained with self-supervision to produce general-purpose dense features, without being taught a fixed notion of an "object". This results in representations that transfer well when the image type changes, and that respond to texture and local structure rather than only to object-like shapes. Conversely, SAM is trained on supervised mask annotations on largely natural imagery. Therefore, it learns strong built-in priors about a well-defined "object's" appearance and its boundaries. Therefore, for each concept, fusing their capabilities intelligently would lead to superior segmentation results. 

Using this understanding, we formulate \method{}, with the following main contributions:
\begin{itemize}
    \item We propose \method{}, a training-free framework that fuses a DINO-based region expert (RSRM) and a SAM-based boundary expert (FSS-SAM3) by assigning one of them to \textit{lead} the segmentation for a given concept. The lead is selected using the segmentation masks computed for the support images: we hold out one support in turn, complete 1. region-led, and 2. boundary-led routes on it, and choose the route whose final output best matches the held-out support's mask. Therefore, the decision reflects the mask that is actually deployed rather than the raw expert prediction.

    \item We introduce a deterministic boundary-correction in the region-led route in which the region expert supplies the base mask and the boundary expert refines its contour within confidence-gated near- and far-pixel bands around the mask's predicted boundary, correcting local shape without allowing the object to be relocated.

    \item We introduce a complementary region-correction in the boundary-led route in which the boundary expert's exterior contour is preserved while the region expert conservatively prunes unsupported components, fills confidently supported holes, and bridges short supported gaps, suppressing implausible regions while keeping accurate boundaries intact.
\end{itemize}

\rev{Across 3,314 matched five-shot episodes from 10 datasets, \method{}
improves equal-dataset foreground mIoU by 5.210 percentage points over RSRM
and has the highest point estimate on seven datasets.}

\section{Related Work}

\subsection{Few-Shot Segmentation}

Few-shot segmentation task is to predict a query mask from a small set of annotated
support images.  Existing methods currently use either prototype-based or affinity-based
matching.  Prototype-based methods summarize the support foreground with one
or more feature vectors and compare them with query features.  This is
efficient, but compressing all support foreground features into a
small number of vectors can hide differences between object parts and boundary
regions.  Affinity-based methods preserve more local information by matching support and query
locations directly.  HSNet, for example, combines correlations from several
feature levels to obtain the query mask \cite{hsnet}.

Recent methods reuse representations from large vision models to improve semantic matching and reduce task-specific training.  However, standard few-shot segmentation generally assumes that training and testing
images come from similar domains. A method designed for natural photographs
might perform poorly on medical scans, satellite images, or industrial
surfaces.

\subsection{Cross-Domain Few-Shot Segmentation}

Cross-domain few-shot segmentation removes the same-domain assumption.  The
support and query images come from a target domain that can differ substantially
from the data used to train or pretrain the model.  PATNet studies how to
transfer useful representations under this domain shift \cite{patnet}.  Other
methods improve support representations, transform features into a more
domain-stable space, or adapt selected model components using the limited
target supports.

Vision foundation models provide stronger general-purpose features for this
setting.  HERA selects and adapts hierarchical features using the support
examples \cite{hera}.  RSRM uses frozen DINOv3 features with semantic
re-fusion, support enhancement, and multi-scale prototype matching
\cite{rsrm,dinov3}.  These methods primarily improve how the target concept is
represented and matched across domains.  In contrast, \method{} studies how
two complete frozen segmentation methods with complementary errors can be assigned
different roles using only the labeled supports.

\subsection{Training-Free Few-Shot Segmentation}

Training-free methods avoid fitting a new segmentation network for the target
task.  Matcher uses frozen foundation-model features to locate related query
regions and then uses SAM to generate the mask \cite{matcher}.  GF-SAM organizes
support--query correspondences as a graph before producing semantic prompts
\cite{gfsam}.  PerSAM personalizes SAM from an annotated reference image
without updating the full model \cite{persam}.

FSS-SAM3 uses labeled visual examples as an in-context prompt to SAM3
\cite{fsssam3,sam3}.  Promptable segmentation models can produce detailed
object boundaries, but a sharp boundary does not ensure that the intended
semantic object has been selected.  This distinction is important under a
large domain shift, where semantic matching and mask generation can fail in
different ways.

\subsection{Boundary Refinement and Expert Combination}

Boundary-refinement methods improve the contour of an existing prediction.
PointRend revisits uncertain locations to produce finer boundaries, while
SegFix learns corrections for predicted boundary pixels
\cite{pointrend,segfix}.  Both approaches refine the output of a single
segmentation model.

Our approach is unique because the chosen segmentation methods have complementary
errors.  One may locate the correct semantic region but produce an imprecise
boundary, while another may generate a clean contour around the wrong object.
Rather than training an additional fusion model or averaging the predictions,
\method{} keeps parameters for both segmentation methods frozen and uses the labeled supports to select
between two fixed correction routes.  The route definitions and support-based
selection procedure are described in Sec.~\ref{sec:method}.

\section{Support-Calibrated Role Fusion}
\label{sec:method}

For segmenting a given concept, \method{} combines a region expert and a boundary expert by letting one of them lead, and letting the other correct it in a confidence-bounded manner. The lead is chosen only using the support images' masks, ensuring that no query annotation ever enters routing or correction. We first provide the problem formulation and definitions of symbols (Sec.~\ref{subsec:setup}), then describe how the lead route is selected on the supports (Sec.~\ref{subsec:calib}), and finally
explain the region-led and boundary-led corrections (Sec.~\ref{subsec:rled}, \ref{subsec:bled}).

\subsection{Problem Setting and Expert Roles}
\label{subsec:setup}

For a concept $c$, let $\mathcal{S}_c = \{(I_i, Y_i)\}_{i=1}^{k}$ be its set of $k$ annotated supports and $I_q$ an unlabeled query image. We use two experts with complementary roles. A \emph{region expert} $R$ locates the correct object, i.e., it produces a binary mask $M^R \in \{0,1\}^{h\times w}$ together with a foreground-minus-background margin $m \in \mathbb{R}^{h\times w}$. $R$ computes confidence scores per pixel $p$ for foreground and background. $m$ is the difference between the foreground confidence and the background confidence. If $m(p)$ is a large positive/negative value, it indicates a confident foreground/background, respectively. $m(p) \sim 0$ indicates low confidence on whether the pixel belongs to the foreground or background.  A \emph{boundary expert} $B$ produces an accurate contour as a binary mask $M^B$. Our concrete instantiation uses RSRM as $R$ and FSS-SAM3 as $B$, and all masks are evaluated on a common $400\times400$ grid.

\method{} has two routes (R-led, and B-led) to compare to achieve an accurate segmentation mask for concept $c$. Both routes rely on how confident $R$ is at each pixel. Therefore, we normalize its margin for the query image without moving its zero,
\begin{equation}
s(p) = \frac{m(p)}
{\max\!\left(\sqrt{|\Omega|^{-1}\textstyle\sum_{u\in\Omega} m(u)^2},\ 10^{-4}\right)},
\label{eq:conf}
\end{equation}
where $\Omega$ is the set of all pixels in $m$. A positive $s$ supports foreground and a negative $s$ supports background; the denominator rescales the margin so that a single confidence threshold has a consistent meaning across images in both routes. We use $\tau = 0.5$ as the confidence threshold on $s(p)$ in both routes.

\subsection{Selecting the Lead Route on the Supports}
\label{subsec:calib}

\begin{revision}
For $k\geq 2$, we compare the two completed routes, each comprising a base
mask and its correction. For fold $i$, we hold out $(I_i,Y_i)$ and predict it from the
remaining $k-1$ supports. Let $\widehat{M}_i^a$ be the completed prediction
for route $a\in\{R,B\}$. The fold score and number of wins are
\begin{align}
j_{ia} &= \operatorname{IoU}(\widehat{M}_i^a,Y_i),
  \label{eq:foldscore}\\
w_a &= \sum_{i=1}^{k}\one\!\left[j_{ia}>j_{i\bar{a}}\right],
  \label{eq:wins}
\end{align}
where $\bar{a}$ is the other route and $\one[\cdot]$ equals 1
when its condition holds and 0 otherwise. The route with more wins leads. Tied
folds count for neither route. If $w_R=w_B$, we pool the true positives,
false positives, and false negatives across folds and select the route
with the higher pooled IoU. An exact pooled tie defaults to the R-led
route. The selected route is fixed for all queries sharing that routing
unit. Section~\ref{subsec:shotbenchmark} describes the one-shot variant.
\end{revision}

\subsection{Region-Led Route and Contour Correction}
\label{subsec:rled}

\begin{revision}
When $R$ leads, its mask $M^R$ supplies the base prediction. On the
$400\times400$ grid, we use a discrete elliptical structuring element of
size $(2r+1)\times(2r+1)$. Write $\delta_r$ for dilation and $\varepsilon_r$
for erosion with this element. The boundary band is
$\mathcal{D}_r(M^R)=\delta_r(M^R)\setminus\varepsilon_r(M^R)$. Let $\mathcal{C}_{\geq 4}(M^B)$ be the union of the 8-connected foreground components of $M^B$ with at least four pixels. Treating masks as foreground
sets, eligible disagreements are
\begin{equation}
\begin{split}
G_r = {} &(M^R\triangle M^B)\cap\mathcal{D}_r(M^R)\\
         &\cap\delta_r\!\left(\mathcal{C}_{\geq 4}(M^B)\right),
\end{split}
\label{eq:eligible}
\end{equation}
where $\triangle$ denotes symmetric difference. This also requires an
edit to lie near a qualifying boundary-model component. The edit set is
\begin{equation}
E_R=G_8\cup\{p\in G_{32}:\lvert s(p)\rvert\leq \tau\}.
\label{eq:editset}
\end{equation}
Pixels in $G_8$ take $B$'s label at any region confidence. Additional
pixels in $G_{32}$ take $B$'s label only if the normalized region margin
satisfies $|s(p)|\leq\tau=0.5$. Every other pixel keeps $R$'s label,
including pixels outside the 32-pixel band. The rule applies to foreground
additions and removals. The completed prediction is
\begin{equation}
\widehat{M}^R(p)=
\begin{cases}
M^B(p),& p\in E_R,\\
M^R(p),& p\notin E_R.
\end{cases}
\label{eq:rled}
\end{equation}
\end{revision}

\subsection{Boundary-Led Route and Region Correction}
\label{subsec:bled}

When $B$ leads, we keep its outline $M^{B}$ and let $R$ act as a structural guard, through three conservative operations applied in turn. Each one either makes a small, well-supported repair but ensure that $B$'s exterior is left intact wherever it is already reliable.

\textbf{Pruning invented regions:} $B$ may sometimes outline a region that is not present. We delete a connected piece $C$ (defined by the pixel connectivity specified in Sec.~\ref{subsec:rled}) within $M^{B}$ only when $M^R$ never overlaps it, and almost all of it is confidently background according to $R$,

\begin{equation}
C \cap M^{R} = \varnothing,
\qquad
|C|^{-1}\!\sum_{p\in C}\rev{\one}\!\left[\, s(p) \le -\tau \,\right] \ge \rho,
\label{eq:prune}
\end{equation}

with $\rho = 0.90$, and $|C|$ denoting the number of pixels in $C$. The first condition checks if $M^R$ has any overlap with $C$. The second condition checks that at least $90\%$ of $C$'s pixels are confidently segmented as background in $M^R$. \rev{A component is removed only when both conditions hold.}

\textbf{Filling supported holes:} $B$ may incorrectly leave a gap inside a real object's mask. Let $H$ denote an enclosed hole. i.e., a background region fully surrounded by the mask and not touching the image border. We fill $H$ only when at least half of it is confidently segmented as foreground in $M^R$ as,

\begin{equation}
|H|^{-1}\!\sum_{p\in H}\rev{\one}\!\left[\, M^{R}(p)=1 \ \wedge\ s(p) \ge \tau \,\right] \ge 0.5.
\label{eq:fill}
\end{equation}

\textbf{Bridging supported fragments.} $B$ sometimes splits a single object into two. We propose short connectors between nearby regions in $M^B$ (\rev{a morphological closing with an elliptical kernel of radius eight pixels}), but keep a proposed pixel only if it is confidently segmented as foreground ($M^{R}(p)=1$ and $s(p)\ge\tau$), and only when \rev{the resulting component joins at least two components of the current mask}. The route therefore cannot conjure an object that $B$ missed entirely. It only reconnects fragments $R$ confidently supports.

Applying the three operations to $M^{B}$ gives the completed route $\widehat{M}^{B}$. In short: keep $B$'s clean outline, and let $R$ only delete regions it flatly rejects, fill gaps it positively supports, and rejoin pieces it confidently connects.

\section{Experiments}
\label{sec:experiments}

\begin{figure*}[!t]
    \centering
    \includegraphics[
        width=\textwidth,
        height=0.42\textheight,
        keepaspectratio
    ]{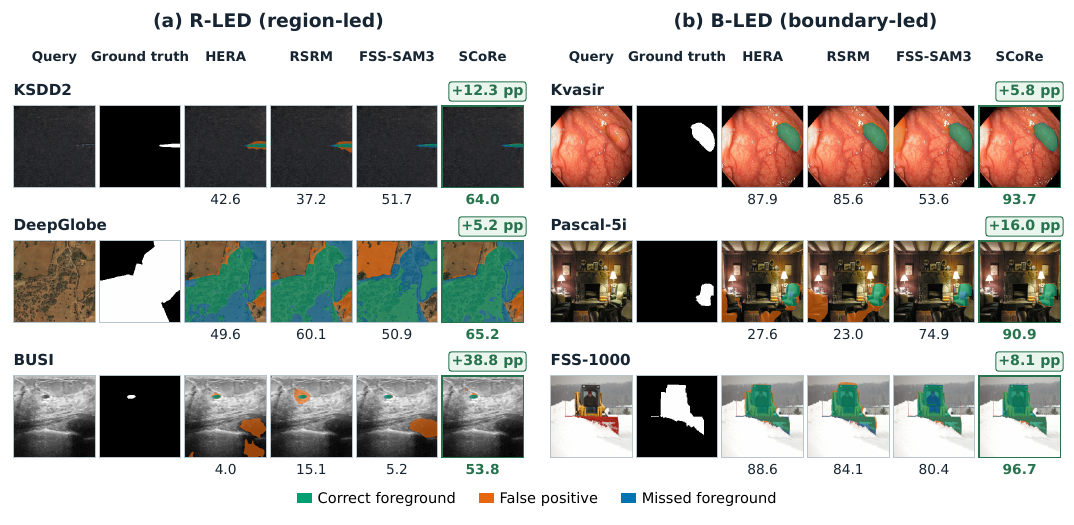}
    \caption{Qualitative results showing how \method{} combines the two
experts. In (a) R-LED, RSRM identifies the target region and FSS-SAM3 refines
its boundaries. In (b) B-LED, FSS-SAM3 provides the contours and RSRM helps
remove false regions and recover missing foreground. Green marks correct
foreground, orange false positives, and blue missed foreground. Scores
below the masks show IoU (\%); the green boxes above the
\method{} predictions show gains over the best displayed baseline
in percentage points. }
    % \caption{Qualitative comparison of R-LED (left) and B-LED (right).
    % Per-image IoU is shown below each prediction. Gains over the strongest
    % displayed individual method are reported in percentage points beside
    % each dataset name.}
    \label{fig:qualitative}
\end{figure*}

\subsection{Datasets and Implementation Details}

\paragraph{Datasets.}
We evaluate on ten datasets covering medical, industrial, natural, aerial, and
road scenes: Chest X-ray \cite{lung}, KSDD2 \cite{ksdd2}, ISIC 2018
\cite{isic}, Oxford-IIIT Pet \cite{pet}, DeepGlobe \cite{deepglobe}, Kvasir-SEG
\cite{kvasir}, BUSI \cite{busi}, \rev{Pothole~\cite{potholemix}}, Pascal-5i \cite{pascal}, and
FSS-1000 \cite{fss1000}.  \rev{The 3,314 five-shot episodes comprise 3,284 foreground and 30 normal KSDD2
episodes. An episode pairs a query image with a support bank. The 177 routing
units are classes, breeds, support banks, or single foreground concepts.}

% Routes are selected for the smallest semantic group defined by each dataset.
% This gives 177 routing units: semantic classes in ISIC, DeepGlobe, Pascal-5i,
% and FSS-1000; breeds in Oxford Pet; three support groups in Kvasir, BUSI, and
% Pothole; and one foreground concept in Chest X-ray and KSDD2.  Therefore,
% different classes within the same dataset may select different routes.

To evaluate the effect of the number of support images on the output, $K\in\{2,5,8\}$ is evaluated on \rev{a separate set of 2,894 matched episodes}.  The two-shot setting uses two images from the five-shot support set.
The eight-shot setting adds three positive examples using the same fixed rule
for every method.  Images used as additional supports are excluded from the
query set, which keeps the comparison across support counts fair.

\paragraph{Compared methods}
We compare RSRM \cite{rsrm}, FSS-SAM3 \cite{fsssam3}, HERA \cite{hera}, Matcher \cite{matcher}, GF-SAM \cite{gfsam}, PerSAM \cite{persam}, and \method{}. 
Matcher and GF-SAM use DINOv2-B/14 to find matching regions between the
support and query images, and SAM-B to generate the query mask. PerSAM-5S
averages the features from the labeled object regions in the five support
images, giving each image equal weight, and uses them to guide the standard
SAM-B prediction procedure. RSRM, HERA, and FSS-SAM3 use their original
inference procedures for the corresponding number of support images.  Only RSRM and FSS-SAM3 are used by \method{}; the other
methods are independent baselines.

\paragraph{Implementation details}
All methods use the same query and support images, and all masks are evaluated
at $400\times400$ resolution.  For $K>1$, each support image is held out once
and predicted from the remaining $K-1$ supports.  The route with more held-out
wins is selected, with pooled IoU used to break a tie as described in
Sec.~\ref{sec:method}.  FSS-SAM3 follows its standard prompt layout for $K=2$
and $K=5$.  For $K=8$, four examples are placed along the top and four along the
left of the prompt canvas.
On the same subset of 2,894 query episodes, \method{} achieves mean mIoU
scores of 68.641\%, 73.118\%, and 72.525\% for $K=2$, $K=5$, and $K=8$,
respectively. Five supports give the highest observed score, while increasing
to eight supports provides no statistically clear improvement. This shows that more support images do not always produce a better route decision.

\paragraph{Evaluation metric}
Following standard CD-FSS evaluation, we report foreground mean intersection
over union (mIoU).  We first pool true positives, false positives, and false
negatives within each class or breed, and then average classes within each
dataset.  The final mean gives equal weight to every dataset. \rev{Foreground mIoU excludes the 30 normal KSDD2 episodes.}

We estimate 95\% confidence intervals by resampling test queries within each class or breed and recalculating the mIoU differences between methods \cite{bootstrap}. Repeated evaluations of the same query image are sampled together, and all methods use the same sampled queries. The support images and selected routes stay fixed during resampling.
\rev{We use 2,000 paired draws per dataset and 10,000 for the equal-dataset mean.}

\subsection{Comparison with Existing Methods}

\begin{table*}[!t]
\caption{\rev{Matched five-shot foreground mIoU (\%). Bold and underline
denote the best and second-best method. R/B counts region-led and boundary-led
routing units. $\Delta$ is the difference against the strongest standalone
method in each row, with a paired 95\% CI, in percentage points.}}
\label{tab:main}
\centering
\scriptsize
\setlength{\tabcolsep}{4pt}
\renewcommand{\arraystretch}{1.08}
\begin{tabular}{lrrrrrrrrrl}
\toprule
Dataset & \rev{Episodes} & \rev{R/B} & RSRM & FSS-SAM3 & HERA & Matcher & GF-SAM & PerSAM-5S & \method{} & \rev{$\Delta$ [95\% CI]} \\
\midrule
Chest X-ray & \rev{60} & 1/0 & \underline{89.104} & 86.107 & 88.671 & 79.990 & 37.207 & 31.258 & \textbf{90.092} & \rev{$+0.988\;[+0.311,+1.621]$} \\
KSDD2 & \rev{60} & 1/0 & 64.386 & 51.844 & \underline{65.867} & 13.339 & 27.966 & 5.766 & \textbf{74.327} & \rev{$+8.461\;[+2.840,+15.016]$} \\
ISIC 2018 & \rev{60} & 3/0 & \underline{68.539} & 67.433 & 66.863 & 38.719 & 43.516 & 17.761 & \textbf{73.374} & \rev{$+4.835\;[+3.081,+7.169]$} \\
Oxford Pet & \rev{74} & 6/31 & 76.051 & \textbf{82.744} & 75.592 & 77.217 & 79.572 & 67.277 & \underline{82.702} & \rev{$-0.042\;[-0.102,-0.00029]$} \\
DeepGlobe & \rev{60} & 6/0 & \underline{57.981} & 48.242 & 56.724 & 53.429 & 53.438 & 41.897 & \textbf{58.413} & \rev{$+0.432\;[-1.066,+1.908]$} \\
Kvasir & \rev{600} & 2/1 & \underline{72.267} & 71.102 & 68.750 & 35.722 & 29.731 & 18.950 & \textbf{72.870} & \rev{$+0.603\;[-1.264,+2.483]$} \\
BUSI & \rev{450} & 3/0 & \underline{40.913} & 30.909 & 34.600 & 13.686 & 12.417 & 8.191 & \textbf{44.104} & \rev{$+3.191\;[+2.018,+4.538]$} \\
Pothole & \rev{450} & 3/0 & \underline{53.654} & 33.407 & 44.199 & 19.746 & 28.565 & 5.601 & \textbf{58.720} & \rev{$+5.067\;[+3.266,+7.166]$} \\
Pascal-5i & \rev{1000} & 5/15 & 70.179 & \textbf{81.756} & 63.811 & 66.732 & 74.899 & 38.901 & \underline{80.951} & \rev{$-0.805\;[-2.048,+0.413]$} \\
FSS-1000 & \rev{500} & 53/47 & 82.205 & \textbf{91.912} & 83.017 & 89.221 & 87.578 & 62.664 & \underline{91.823} & \rev{$-0.089\;[-0.658,+0.574]$} \\
\midrule
\rev{Equal-dataset mean} & \rev{3314} & 83/94 & \underline{67.528} & 64.546 & 64.809 & 48.780 & 47.489 & 29.827 & \textbf{72.738} & \rev{$+5.210\;[+4.618,+5.972]$} \\
\bottomrule
\end{tabular}
\end{table*}

\rev{Table~\ref{tab:main} shows the highest point estimate for \method{} on
seven of ten datasets. The equal-dataset mean exceeds RSRM by 5.210
percentage points (paired 95\% CI [4.618, 5.972]). FSS-SAM3 leads
on Oxford Pet, Pascal-5i, and FSS-1000.}

% \subsection{Support-Count Ablation}

% \begin{table}[!ht]
% \caption{Support-count ablation on 2,894 shared query episodes.  R/B gives the
% number of R-LED/B-LED routing units. Column-best and second-best scores are bold and underlined.}
% \label{tab:shot-ablation}
% \centering
% \scriptsize
% \setlength{\tabcolsep}{3.0pt}
% \renewcommand{\arraystretch}{1.06}
% \begin{tabular}{crrrr}
% \toprule
% $K$ & RSRM & FSS-SAM3 & \method{} & R/B \\
% \midrule
% 2 & 65.943 & 56.762 & 68.641 & 67/110 \\
% 5 & \underline{68.041} & \underline{64.474} & \textbf{73.118} & 83/94 \\
% 8 & \textbf{68.819} & \textbf{64.759} & \underline{72.525} & 56/121 \\
% \bottomrule
% \end{tabular}
% \end{table}

% Table~\ref{tab:shot-ablation} studies the effect of the number of support
% images.  All three settings use the same 2,894 queries.  This is why the
% five-shot value in Table~\ref{tab:shot-ablation} differs from the value in
% Table~\ref{tab:main}, which uses all 3,314 queries.

% Increasing the support count from two to five improves \method{} by 4.477
% points, with a 95\% confidence interval of $[+3.493,+5.469]$.  Increasing it
% from five to eight changes the score by $-0.593$ points, with an interval of
% $[-1.447,+0.255]$.  Thus, the improvement from two to five supports is clear,
% while the difference between five and eight supports is not.  The raw experts
% improve slightly at eight supports, but the route assignments also change.  This
% shows that more support images do not always produce a better route decision.

\subsection{\rev{Benchmark and Fusion Analysis}}
\label{subsec:shotbenchmark}

\rev{Table~\ref{tab:shotbench} reports single-seed results on four
standard CD-FSS datasets~\cite{patnet}, with matched episodes at each
shot count. RSRM, HERA, and FROST~\cite{park2026frost} share DINOv3
ViT-L/16, and FSS-SAM3 uses SAM3. For one-shot inference, we use the
first of the five supports. To calibrate the route, we predict each of
its three views (original, horizontal mirror, and $1.25\times$ contrast)
from the other two.}

\begin{table}[!t]
\caption{\rev{Single-seed mIoU (\%), shown as 1-shot/5-shot.
Each setting uses 1,200 episodes per dataset (2,400 for FSS-1000).
The mean weights datasets equally. Bold marks the best score per shot.}}
\label{tab:shotbench}
\centering
\scriptsize
\setlength{\tabcolsep}{2.0pt}
\revisionink
\begin{tabular}{lrrrrr}
\toprule
Dataset & RSRM & FSS-SAM3 & HERA & FROST & \method{} \\
\midrule
DeepGlobe   & 48.32/59.04 & 30.93/46.17 & 43.18/59.62 & \textbf{50.53}/\textbf{59.84} & 48.12/58.15 \\
ISIC 2018   & 57.62/65.67 & \textbf{64.41}/70.21 & 52.42/61.78 & 55.50/66.68 & 61.98/\textbf{71.36} \\
Chest X-ray & 87.57/88.92 & 87.02/82.58 & 75.70/87.46 & 84.69/86.06 & \textbf{90.15}/\textbf{88.98} \\
FSS-1000    & 80.68/83.06 & 88.06/91.24 & 80.15/82.42 & 82.92/86.26 & \textbf{88.87}/\textbf{91.36} \\
\midrule
Mean        & 68.55/74.17 & 67.60/72.55 & 62.86/72.82 & 68.41/74.71 & \textbf{72.28}/\textbf{77.46} \\
\bottomrule
\end{tabular}
\end{table}

\begin{revision}
In the ten-dataset ablation study, fixed R-led reaches 72.553\%
mIoU, 5.025 percentage points above raw RSRM. Fixed B-led scores
64.747\%, and the query-IoU oracle reaches 74.884\%. Support-based
routing adds 0.185 points over fixed R-led (paired 95\% CI
[\(-0.010\), \(+0.386\)]). With routing recalibrated on the original
support folds after each removal, dropping the outer extension
causes the largest decrease (2.975 points). The other decreases are
0.058 (inner override), 0.012 (pruning), 0.001 (hole filling), and
0.007 (bridging).

The 8- and 32-pixel radii come from exploratory runs on DeepGlobe,
ISIC, and Chest X-ray. We set thresholds retrospectively and used
them across datasets. Varying one setting at a time, inner radii
4/12 give 72.680/72.676\% mIoU and outer radii 24/40 give
72.515/72.824\%. Cutoffs of 0.25/0.75 give 70.879/72.501\% for the
outer band and 72.709/72.721\% for B-led.

At a three-pixel tolerance, FSS-SAM3 scores 49.36\% mean BF1 versus
29.50\% for RSRM and 50.50\% for \method{}. In a five-shot pilot
with 12 matched episodes per dataset, we replace RSRM with FROST
and use the same correction rules. Fusion reaches 75.97\% mean
foreground mIoU versus 73.92\% for FROST and 68.03\% for FSS-SAM3.
It beats both on DeepGlobe and FSS-1000. On the same episodes,
RSRM--HERA trails RSRM by 2.08 percentage points.

On an RTX 3090, the RSRM--FSS-SAM3 pair takes 0.60/1.36~s per
query (median, batch size one, cached route) and 1.91/4.58~s to
calibrate each support bank at one/five shots.
\end{revision}

% We also examine the 30 normal KSDD2 images.  FSS-SAM3 predicts foreground on
% 63.3\% of them, with a mean foreground area of 0.005\%.  RSRM and \method{}
% predict foreground on 93.3\%, with a mean area of 0.429\%.  Therefore,
% \method{} should be viewed as a foreground segmentation method, not as an
% anomaly detector.

% \begin{figure}[!htbp]
%     \centering
%     % \setlength{\abovecaptionskip}{0pt}
%     % \setlength{\belowcaptionskip}{0pt}
%     \includegraphics[width=\columnwidth]{qualitative_successes_single_column.pdf}
%     \caption{Qualitative comparison on representative examples. The IoU of each prediction is shown below its mask. The rightmost column reports SCoRe's improvement over the strongest individual expert in percentage points (pp), together with the support-selected route.}
%     % \caption{Qualitative comparison across the evaluated datasets.}
%     \label{fig:qualitative}
% \end{figure}

\subsection{Qualitative Results}

Figure~\ref{fig:qualitative} presents examples from \rev{six datasets} showing boundary refinement,
removal of incorrect structures, and confidence-guided correction done by \method{}. It shows how the two routes use complementary expert predictions.  In the R-LED examples, RSRM generally identifies the main object region, while FSS-SAM3 contributes useful local boundary information.  The
KSDD2 and BUSI examples show the removal of false-positive regions around small
targets, while \rev{the DeepGlobe example shows a local correction that preserves the overall region.}  The Kvasir, Pascal-5i, and FSS-1000 examples demonstrate B-LED route in which
FSS-SAM3 provides the base mask, while RSRM evidence supports conservative
structural corrections. 

% Figure~\ref{fig:qualitative} presents examples from the seven datasets showing boundary refinement,
% removal of incorrect structures, and confidence-guided correction done by \method{}. For each dataset, we show the positive query with the largest per-image gain
% over the strongest individual expert. This selection is performed only for
% visualization after all predictions are fixed; query annotations are not used
% for route selection or mask generation.

\FloatBarrier

% \begin{figure*}[!t]
% \centering
% \includegraphics[width=0.82\textwidth]{qualitative_successes.pdf}
% \caption{Examples from the seven datasets led by \method{}.  Green denotes
% correct foreground, orange denotes false positives, and blue denotes missed
% foreground.  The lower-left value is the image IoU.  The upper-right value is
% the improvement over the strongest displayed individual method, followed by
% the selected route.}
% \label{fig:success}
% \end{figure*}

% \FloatBarrier

% \input{5_Results}
\section{Conclusion}

We presented \method{}, a training-free framework that assigns complementary
roles to frozen few-shot segmentation methods.  Five-way leave-one-out evaluation uses only
the labeled supports to choose a completed R-LED or B-LED route for each routing
unit. RSRM provides region information and FSS-SAM3 provides boundary information.  The
route is selected from the labeled supports using a fixed rule, without an
additional learned model or query annotations.

\rev{On ten matched five-shot datasets, \method{} reaches 72.738\% foreground
mIoU versus 67.528\% for RSRM and has the highest point estimate on seven
datasets.}  A matched support-count study shows a clear gain from two to five
supports, but no reliable gain from five to eight.  The method is most effective
when the two experts make complementary region and boundary errors.  The lower
scores on Oxford Pet, Pascal-5i, and FSS-1000 show that route selection does not
guarantee the best result on every dataset.

The main limitation is the reliability of support-based calibration.  The
leave-one-out folds are correlated, and a small support set may not represent
difficult query appearances.  The selected route can consequently be suboptimal for some queries. In addition, both experts must run during inference, and neither correction route guarantees that every edit will improve the base prediction. Future work will explore more reliable route selection, safer correction rules, and ways to avoid running both experts when one is sufficient.

\bibliographystyle{IEEEtran}
\begingroup
\ifrevisionmarks
  \let\scorebibitem\bibitem
  \renewcommand{\bibitem}[1]{%
    \color{black}\scorebibitem{#1}%
    \ifcsname scorebibrev@#1\endcsname\revisionink\fi}
  \expandafter\def\csname scorebibrev@potholemix\endcsname{}
  \expandafter\def\csname scorebibrev@park2026frost\endcsname{}
\fi
\bibliography{References_new}
\endgroup

\end{document}